# Big-Data-Driven and AI-Based Framework to Enable Personalization in Wireless Networks

Rawan Alkurd, Ibrahim Abualhaol, and Halim Yanikomeroglu

*The authors propose the utilization of AI, big data analytics, and real-time non-intrusive user feedback in order to enable the personalization of wireless networks. Based on each user's actual QoS requirements and context, a multi-objective formulation enables the network to micro-manage and optimize the provided QoS and user satisfaction levels simultaneously.*

## Abstract

Current communication networks use design methodologies that prevent the realization of maximum network efficiency. In the first place, while users' perception of satisfactory service diverges widely, current networks are designed to be a "universal fit," where they are generally over-engineered to deliver services appealing to all types of users. Also, current networks lack user-level data cognitive intelligence that would enable fast personalized network decisions and actions through automation. Thus, in this article, we propose the utilization of AI, big data analytics, and real-time non-intrusive user feedback in order to enable the personalization of wireless networks. Based on each user's actual QoS requirements and context, a multi-objective formulation enables the network to micro-manage and optimize the provided QoS and user satisfaction levels simultaneously. Moreover, in order to enable user feedback tracking and measurement, we propose a user satisfaction model based on the zone of tolerance concept. Furthermore, we propose a big-data-driven and AI-based personalization framework to integrate personalization into wireless networks. Finally, we implement a personalized network prototype to demonstrate the proposed personalization concept and its potential benefits through a case study. The case study shows how personalization can be realized to enable the efficient optimization of network resources such that certain requirement levels of user satisfaction and revenue in the form of saved resources are achieved.

## Introduction

Continuously advancing technology has contributed to a surge in data traffic, making user satisfaction the cardinal competitive advantage for all service providers. Besides user satisfaction, service providers try to make the most of the scarce resources available. In order to meet these objectives simultaneously, more agile, intelligent, and flexible networks are required. Such networks should be capable of micro-managing resources in a way that meets each user's expectations of the network while using a minimum amount of resources. This micro-management of network resources has ushered in the concept of wireless network personalization. Personalized networks optimize two correlated and contradicting objectives in real time: user satisfaction and resource utilization. Naturally, wireless networks produce colossal amounts of data, and most of these data are in real time. A system that is capable of digesting these data to create relevant and meaningful decisions in real time at the user level using machine learning (ML) and big data analytics is the ultimate solution to meeting the aforementioned objectives simultaneously. We refer to such a system as a big-data-driven and artificial intelligence (AI)-based personalized wireless network. Enabling wireless network awareness of context data and user feedback data, and consequently enabling wireless network personalization will bring substantial benefits to both users and service providers [1, 2].

The utilization of context data to optimize and enhance the performance of wireless networks has been proposed in [3]. Although context data plays an important role in improving the performance of wireless networks, making network decisions based on both context information and user feedback will provide service providers with tangible data that can be utilized to make further optimized and personalized decisions. User feedback collection can be performed in real time or offline in a number of ways that can be either intrusive (i.e., require human input), such as surveys and feedback boxes, or non-intrusive, which is achieved through ML and AI. In wireless networks, the utilization of user feedback from intrusive methods is discussed in [4, 5]. The authors in [4] propose an approach called "user-in-the-loop" that utilizes real-time feedback to integrate spatial demand control to wireless networks where users are motivated to move to less congested areas. The authors in [5] propose a data-guided resource allocation approach where offline feedback data (e.g., network measurements and user complaints) are employed to improve the average user experience. The utilization of non-intrusive user feedback in wireless communication networks is not a common discussion topic in the literature and is limited to but a few applications [6]. Nevertheless, non-intrusive user feedback has been proposed to make relevant automated decisions in many applications, such as cloud gaming and healthcare [7, 8].

The detection of non-intrusive user feedback is widely discussed in the computational intelligence literature [9, 10]. There are several advantages to employing non-intrusive feedback collection methods, over intrusive methods, in wireless networks. Generally, the intrusive feedback collection methods do not represent all users because the majority of users would not complain; they just change their provider. Furthermore, while users' needs and expectations change rapidly in wireless networks, non-intrusive feedback collec-tion methods enable more frequent feedback data collection, which consequently increases the accuracy and relevance of network decisions.

This article proposes the utilization of context data along with non-intrusive user feedback data to personalize wireless networks. While the proposed personalization concept could potentially be applied to all wireless networks, we focus in this article specifically on wireless cellular networks as a use case. We introduce the concept of wireless network personalization through addressing the following four important questions:
- *Q1:* Why does personalization matter?
- *Q2:* What is the difference between personalization and network slicing?
- *Q3:* How is user satisfaction measured in wireless networks?
- *Q4:* How is personalization integrated into wireless networks?

First, to answer Q1, we start by discussing the benefits of personalized wireless networks. For Q2, we narrow down our discussion to address personalization for cellular networks and how it differs from 5G network slicing. For Q3, we introduce the notion of the zone of tolerance, which we use to propose the user satisfaction model in wireless networks. Then we illustrate how the proposed user satisfaction model can be used in the context of personalized wireless networks. To answer Q4, we propose a big-data-driven and AI-based framework to enable personalization in wireless networks. Finally, we illustrate the proposed wireless network personalization concept and the associated benefits through a prototyped case study.

## Why Does Personalization Matter?

Most service providers are scrambling to increase the average revenue per user and to reduce subscriber churn. Therefore, there is a tremendous need to efficiently utilize scarce resources in order to achieve the ultimate balance between user satisfaction and profit. Personalization is indispensable to achieving this balance and delivering services profitably to users in a win-win setting. In addition, there are several other benefits of integrating personalization into wireless networks, including the following.

### Not Everyone Fits the Mold

Typically, service providers invest in their networks to acquire more subscribers and increase their revenue. Therefore, they always seek to provide their subscribers with the best service quality. Service quality is defined as a comparison between subscribers' expectations and service performance [11]. Current networks are designed mostly to be a "universal fit," where service providers deliver services with a quality appeal to all types of users. However, user expectations of service quality are not "one size fits all." In practical terms, it is difficult to measure user expectations of service quality for current networks as this depends on numerous dynamic and difficult-to-measure variables. For this reason, service providers over-engineer the delivery of services, and as a result, many users end up getting more resources than would actually satisfy them, while others end up getting less than they expect. The way around this inefficiency is to tailor the network for each user's dynamic and context-dependent needs and expectations. This level of fine-grained network decision optimization will enable service providers to provide personalized, satisfactory services for the majority of users at a minimum cost.

### Act at the Speed of Users

Although users could show some patterns in their long-term activity and satisfaction behavior, the real-time (short-term) user expectations and behavior evolve continuously. In order to make personalized decisions and actions dynamically, networks need to decide and act at the speed of users. Personalized networks employ ML and big data analytics, which make real-time network decisions and actions possible through automation. Automation can be achieved by analyzing the enormous amounts of data produced by networks to identify relevant patterns and thereby predict context-dependent user needs and expectations.

### Pricing Paradigms

Currently, tariffs are differentiated based on usage. Wireless network personalization will fit well with the pricing paradigms differentiated according to quality of service (QoS) and user satisfaction levels. In addition, not only are QoS and user-satisfaction-based pricing paradigms fairer and more attractive to users, but they will also create new business models and revenue opportunities for service providers. Based on the new business models, service providers can trade off complexity and the required user satisfaction and resource-saving levels in the network.

## Personalization vs. Network Slicing in Wireless Cellular Networks

Network function virtualization (NFV) technology is proposed for 5G to isolate the software and hardware aspects of networks in order to transform network functions from dedicated hardware appliances into software-based applications. Along with NFV, software-defined networks (SDNs) are considered to be enablers for network slicing (NS) in 5G [12]. The concept of NS is proposed to allow operators to provide customized, reliable services with increased efficiency while reducing the capital expenditure and operating expenses of wireless networks. Each slice is associated with a set of resources and quality requirements. What distinguishes 5G NS from the current QoS-based solutions is its ability to provide an end-to-end virtual network tailored to application requirements. In contrast, the personalized networks advocated in this article are not only tailored to application requirements, but also to the dynamic user demands and expectations within the application. In other words, instead of specifying a fixed set of quality requirements, personalized networks are supported with an intelligent big-data-driven layer to decide on the dynamically evolving user requirements. This enables the service providers to provide the necessary service quality required to achieve the target user satisfaction level for each user and any application.

## Modeling User Satisfaction in Wireless Networks

### User Zone of Tolerance Model

Due to the intense competition in the telecommunication industry, user satisfaction is crucial to sustaining a profitable business for any service

> There is a tremendous need to efficiently utilize scarce resources in order to achieve the ultimate balance between user satisfaction and profit. Personalization is indispensable to achieving this balance and delivering services profitably to users in a win-win setting. In addition, there are several other benefits of integrating personalization into wireless networks.

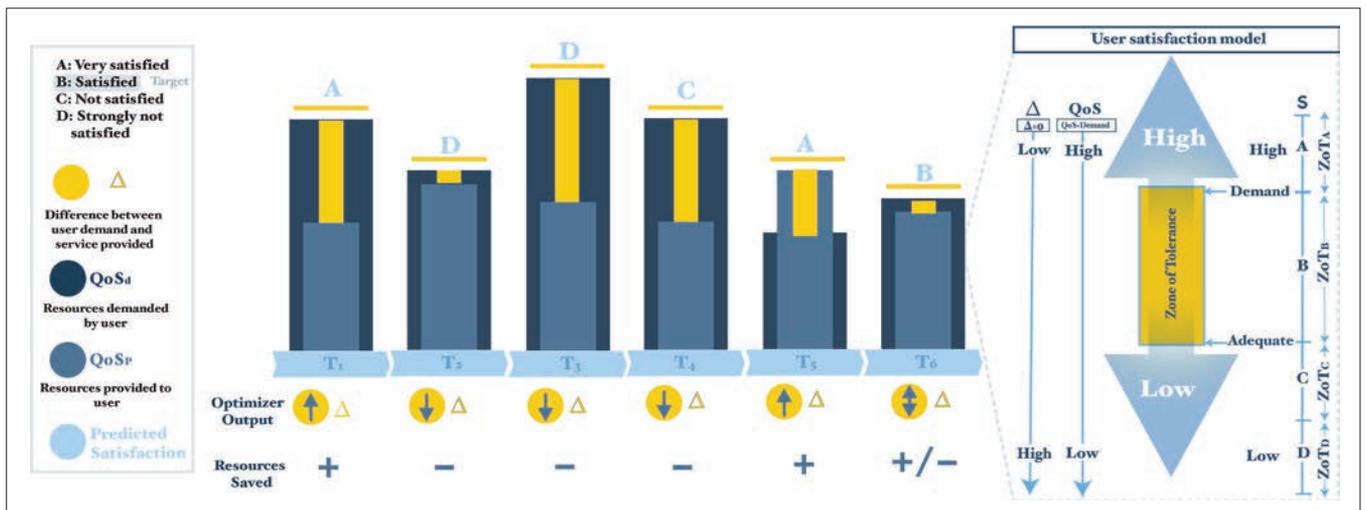

**Figure 1.** User satisfaction (S) model and a visualization of an example illustrating the relationship between the zone of tolerance, $\Delta$, user satisfaction, and personalized network decisions.

provider. Over the long run, service providers that can maintain user satisfaction will obtain and retain more subscribers and increase business growth. "Satisfied customers offer businesses a promise of enhanced revenues and reduced operating costs" (Dutka, 1995). In order to understand customer/user satisfaction, business and marketing studies proposed the model of service quality. This model utilizes the zone of tolerance (ZoT) notion, which is defined as "the range of service performance a customer would consider satisfactory." The model of service quality depicts the ZoT as the service range bounded by desired and adequate levels of service. The desired level of service is defined as "the level of service representing a blend of what customers believe can be and should be provided," whereas the adequate level of service is defined as "the minimum level of service customers are willing to accept."

Drawing on concepts of service quality from business and marketing studies, our model of user satisfaction (S) in wireless networks is shown in Fig. 1. We propose dividing user satisfaction into levels where each level is associated with a certain range of QoS. In Fig. 1, satisfaction is divided into four levels: A, B, C, and D. The division and number of satisfaction levels could vary depending on service providers' preferences. Before proceeding to the detailed discussion of the user satisfaction model, it is worth mentioning that QoS can be a vector with several elements, such as rate, reliability, latency, and jitter. Nonetheless, for simplicity, we assume that QoS is solely defined by rate. Our proposed user satisfaction model encompasses the following five main notions:

- $QoS_d$: the demanded QoS by the user, which represents the maximum QoS associated with the requested service.
- $QoS_p$: the provided QoS by the network.
- $QoS_{a_i}$: the adequate (minimum) QoS required to achieve a satisfaction level of $i$.
- $ZoT_i$: the QoS range that satisfies the user with a satisfaction level of $i$. If satisfaction is measured on a scale of 1–5, for $i < 5$, $ZoT_i$ ranges between $QoS_{a_{i+1}}$ and $QoS_{a_i}$, whereas for $i = 5$, $ZoT_5$ ranges between $QoS_d$ and $QoS_{a_5}$.
- $\Delta$: the difference between the QoS demanded by the user and the QoS provided by the network ($QoS_d - QoS_p$).

As shown in Fig. 1, as $QoS_p$ decreases, $\Delta$ increases, and consequently satisfaction decreases. To keep user satisfaction at a certain level, $QoS_p$ should be within the ZoT associated with the targeted satisfaction level. It is important to note that $QoS_{a_i}$ is what changes from one user to another, which consequently changes the width of $ZoT_i$. Moreover, demand is assumed to be dependent on the application and service type; hence, it is constant for all users requesting service of the same application.

### Zone of Tolerance in the Context of Personalized Networks

Understanding and characterizing ZoT will open the door to micro-managing wireless networks, which will allow operators to personalize their services and design new business models to generate new revenue streams while maintaining user satisfaction. In Fig. 1, we present a simple example to illustrate how ZoT, $\Delta$, and user satisfaction are related. Assume that a service provider is trying to optimize the network such that a certain user in the network has a satisfaction level of B. Furthermore, assume that the service provider is utilizing a big-data-driven and AI personalized network to predict $ZoT_B$ at different time slots ($T_1$ to $T_6$). Based on the predicted $ZoT_B$, the personalized network optimizes $\Delta$ during each time slot. As shown in Fig. 1, at $T_1$, based on the provided $\Delta$, the predicted user satisfaction level is A. In order to reduce satisfaction and consequently save resources, the optimizer, which is part of the personalized network, suggests increasing $\Delta$ ⇑ to save resources (+). At $T_2$, since $ZoT_B$ is continuously changing over time, although the provided $\Delta$ is lower than $\Delta$ at $T_1$, user satisfaction drops to level D. This indicates that this user has a very tight $ZoT_B$ at $T_2$, and consequently the optimizer suggests decreasing $\Delta$ ⇓ further by allocating more resources (−). At $T_6$, the predicted user satisfaction for the provided $\Delta$ is level B, which is the targeted user satisfaction. Therefore, the optimizer suggests keeping $\Delta$ as is ⇕ (+\−).

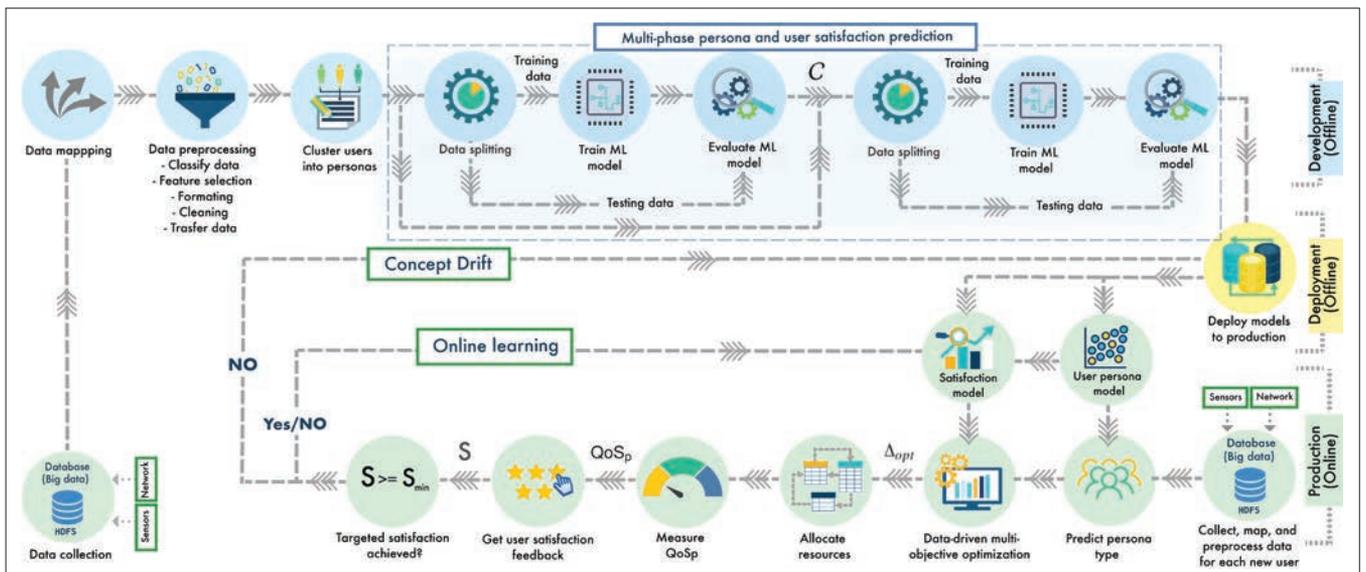

Figure 2. Big-data-driven and AI-based network personalization framework.

## WIRELESS NETWORK PERSONALIZATION: BIG-DATA-DRIVEN AND AI-BASED FRAMEWORK

Data collected from communication networks are massive, complex, and unstructured, and increase in three dimensions: volume, velocity, and veracity. The problem of extracting knowledge from this huge amount of data presents two subproblems: a big data problem and an AI problem. AI is defined as any process that senses the environment and takes actions to maximize the success probability of the targeted goal. Until recently, it has not been feasible to solve such a problem in real time. However, supercomputers and distributed computing technologies are improving rapidly to the point that the use of big data analytics and prediction techniques for practical near-real-time applications are currently possible.

As shown in Fig. 2, the proposed personalization framework collects information from the user environment and the network, predicts user needs and tolerance to service quality, and optimizes resource allocation to minimize cost and maintain certain user satisfaction levels. The proposed framework consists of three stages:

### DEVELOPMENT

The development stage is composed of the following modules, all of which are implemented offline.

**Data Mapping:** Data from different users are mapped to shared space. Mapping user data is an essential step as it enables ML models to capture correlations and inherent patterns. For instance, user location is recorded as global positioning system (GPS) coordinates. However, generally, user satisfaction behavior is actually correlated to a particular type of location (e.g., home) rather than GPS coordinates.

**Data Preprocessing:** Retransform the dataset and extract useful features.

**Cluster Users into Personas:** A group of users who share similar user behavior and satisfaction patterns are referred to as a persona [2]. Associating users with pre-existing user personas will enable networks to provide highly personalized services with a minimal amount of data. Besides, the utilization of federated learning can enable different users sharing the same persona type to learn a shared model while keeping all the training data on their devices. This can reduce the required communication overhead and minimize the privacy concerns associated with transferring user data over the network.

**Multi-Phase Persona and User Satisfaction Prediction:** At this point, the network has access to labeled context data with user persona and satisfaction levels. The processed data is used to build an ML model to predict user satisfaction levels for each user. As shown in Fig. 2, we approach the user satisfaction prediction problem using a two-phase ML model. The first phase is designed to output the persona's probability vector **C**. The second phase digests **C** along with the preprocessed labeled data in order to build a model capable of predicting the user satisfaction levels for new and existing users using a minimum amount of data.

### DEPLOYMENT

At the deployment stage, the output ML model from the previous stage is integrated into the production environment to start making practical decisions based on new data. There are several methods used to deploy ML models. In our framework, to automate the prediction process, the ML model is deployed as an online ML model. Online ML models continue to update and train as more data becomes available to the network.

### PRODUCTION

The production stage is where the network utilizes the trained ML models to achieve network personalization in real time. The production stage is composed of the following modules.

**Data Collection and Preprocessing:** The first step is to continuously collect and preprocess context information from users to predict personalized user satisfaction behavior.

**Data-Driven Multi-Objective Optimization:** The next step is to use the multi-phase ML model trained in the development stage as an input to a multi-ob-jective optimization problem. The optimization problem is formulated to micro-manage and optimize resources and users' satisfaction simultaneously based on each user's QoS requirement and user satisfaction behavior. This optimization problem utilizes the ML model as its fitness function. Moreover, the targeted satisfaction level for each user is decided by the network service provider and fed as an input to the multi-objective optimization problem. The optimization problem outputs the optimum choice of $\Delta$ ($\Delta_{opt}$), which achieves the required satisfaction level using the minimum amount of resources.

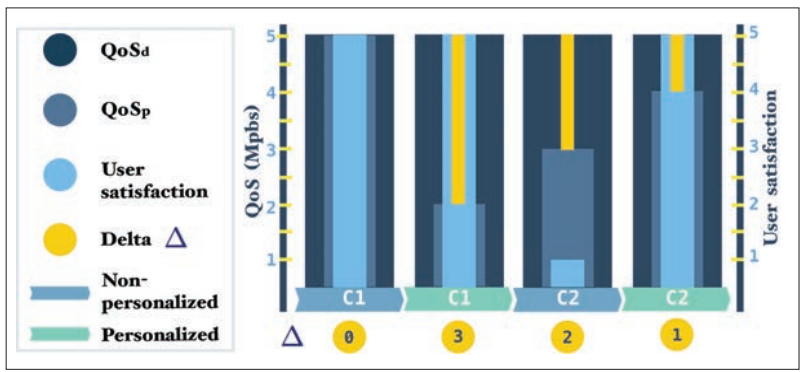

**Figure 3.** $QoS_d$, $QoS_p$, $\Delta$, and the corresponding user satisfaction for contexts $C_1$ and $C_2$ associated with both personalized and non-personalized networks.

**Resource Allocation:** The network utilizes $\Delta_{opt}$ and other network parameters to allocate the best resource blocks (RBs) to achieve the targeted user satisfaction level.

**Measure $QoS_p$ and Get User Satisfaction Feedback:** After allocating resources to each user, the network records the $QoS_p$ along with user feedback (i.e., satisfaction).

**Tuning:** The measured user satisfaction is used to validate the predicted satisfaction levels. If user satisfaction was predicted correctly, the instance is fed to the database. Otherwise, the error is fed to the predictive model to relearn (i.e., concept drift). Relearning is used to improve the predictive model performance and update the model with user behavioral changes that could occur over time.

**Online Learning:** Online learning is used to improve the predictive model proactively. Since network data become available in sequential order, batch learning techniques are not practical for real-time implementation. Instead, online learning techniques can dynamically adapt to new changes or patterns in user behavior and its relation to satisfaction.

In order to assess the practicality of the proposed framework, the steps that need to be performed during the communication session (i.e., online) should be assessed. As shown in Fig. 2, the production process involves relatively fast operations, such as data collection and performing predictions. However, utilizing optimization algorithms to solve data-driven resource allocation problems could potentially be a time-consuming process. Since optimizing resources in near real time is essential for practical systems, meta-heuristics can provide sufficiently good solutions in a relatively short time. In contrast, the development and deployment stages may involve cumbersome, time-consuming processes, such as initial data collection, training, and validation, which require heavy computation. Nonetheless, since they are implemented offline, they should not affect the network proactivity.

## How Can Personalization Potentially Save Resources and Increase User Satisfaction?

In this section, we shall consider a few exemplary instances in order to illustrate how personalization can potentially save resources and improve satisfaction at minimum cost. Figure 3 illustrates two different contexts, $C_1$ and $C_2$. The value of $\Delta$ is illustrated by the length of the yellow bar and is depicted, for each context, in the yellow circles. User satisfaction is measured on a scale of 1 to 5. The targeted user satisfaction for the considered user is assumed to be 5, and the $QoS_d$ is 5 Mb/s.

To begin with, we analyze the data associated with $C_1$. As for the non-personalized network, the allocated $QoS_p$ was 5 Mb/s, and consequently, $\Delta$ will sum up to 0 Mb/s. On the other hand, the personalized network allocated only 2 Mb/s, causing $\Delta$ to rise to 3 Mb/s. The personalized network was able to predict that this user, during $C_1$, would have a relatively large $ZoT_5$ and hence reduce the amount of resources allocated by 3 Mb/s without sacrificing user satisfaction. Accordingly, we can conclude the following: *personalization can potentially save resources during contexts with larger ZoT without sacrificing user satisfaction.*

On the contrary, during $C_2$, the non-personalized network allocated 3 Mb/s, whereas the personalized network allocated 4 Mb/s to reduce $\Delta$ from 2 to 1 Mb/s. As a result, user satisfaction climbed from 1 for the non-personalized network to 5 for the personalized network. Personalization enabled the network to predict that this user, during $C_2$, would have a smaller $ZoT_5$, and hence the minimum required $QoS_p$ is 4 Mb/s. Accordingly, we can conclude the following: *personalization can potentially increase user satisfaction to the desired level using a minimum amount of resources.*

### Prototyping a Personalized Wireless Network

In order to verify the benefits of personalized networks, we implemented the proposed framework as a proof-of-concept case study, which resembles a simplified but realistic network scenario. The prototype was implemented in python where the TensorFlow library was used to build the ML capabilities required for the framework. In this section, we utilized the implemented prototype to study the benefits of integrating personalization into wireless networks. For the purpose of comparison, we implemented two networks, a personalized one and a non-personalized one. The personalized network utilizes the proposed framework to minimize the allocated resources (i.e., maximize $\Delta$) while maintaining user satisfaction levels higher than the targeted minimum satisfaction. On the other hand, since current cellular networks are designed to maximize $QoS_p$, the implemented non-personalized network optimizes the allocated resources to provide services with $QoS_p$ as close as possible to $QoS_d$ (i.e., minimize $\Delta$) for all users.

#### Experimental Setup

**Cellular Network Environment:** Consider a cell within a cellular network that covers Ottawa, Canada. The cell has one eNB and is connected to

22                                                                                                              IEEE Communications Magazine • March 2020 three active users moving within its coverage area. The area of the cell is divided into a $k \times k$ grid. The cellular network environment is simulated using the parameters listed in Table 1. The cellular network operator collects data about the users and stores it in a database. The collected data are of two types, real-time user satisfaction levels as well as context values, such as time, location, and application. Measurements are recorded at each measuring instant. The period between two measuring instances is referred to as a time slot (TS). The operator collects data from the considered users with TS length of 1 s. Furthermore, the amount of resources used for each TS is recorded. In this section, satisfaction values are modeled as discrete values (1–5). Based on the type of service plan provided for the three users, the network operator targets a satisfaction level of 4 ($S_{min}$ = 4).

**Dataset Description:** As shown in Fig. 2, user and network data are crucial requirements for personalized networks. Unfortunately, companies and institutions capable of collecting such data, particularly user data, do not publish them for privacy and confidentiality reasons. The way around this issue is to design and generate synthetic data that is flexible and has realistic characteristics. In [1, 2], we proposed a synthetic dataset design to enable big-data-driven wireless network personalization. The dataset is designed with four distinct user personas and is composed of context features, such as time, location, speed, and application, along with their associated satisfaction labels. The dataset is utilized to build the prototype for the proposed personalized network, and it can be found in a publicly available GitHub repository [13].

## EXPERIMENTAL RESULTS

In this section, for the purpose of comparison, we simulate both personalized and non-personalized networks. The first premise of personalized networks is their ability to minimize the overall utilized resources at each instant (or frame). While resources in wireless networks are miscellaneous, in this article, we confine resources to bandwidth, which is proportional to $QoS_p$ in megabits per second. The amount of saved resources is measured by calculating the difference between the $QoS_p$ provided by the non-personalized network ($QoS_{NP}$) and $QoS_p$ provided by the personalized network ($QoS_{Pr}$) (i.e., $QoS_{NP} - QoS_{Pr}$). In Fig. 4, we plot the total $QoS_{NP}$, $QoS_{Pr}$, $QoS_d$, and $QoS_{NP} - QoS_{Pr}$ for the three users vs. time in hours. As Fig. 4 shows, the total amount of saved resources ($QoS_{NP} - QoS_{Pr}$) fluctuate with time. Essentially, the personalized network achieves the highest resource saving when the non-personalized network attempts to maximize $QoS_p$ while the user has more tolerance to lower $QoS_p$. In this particular scenario, as shown in Fig. 4, the amount of saved resources was always greater than zero, indicating that the personalized network was able to provide service with $QoS_{Pr} \leq QoS_{NP}$; hence, it was able to save more resources (9703.8 Mbits over 24 hours) compared to the non-personalized network. However, depending on $S_{min}$, users' ZoTs, and the amount of resources provided to each user, the personalized network might suggest an increase in the provided resources to certain users (i.e., $QoS_{Pr} > QoS_{NP}$) to push their satisfaction levels above the targeted minimum. Usually, this increase in the provided resources for low-tolerance users is offset by the reduced amount of provided resources for high-tolerance users. In addition, the extra amount of resources suggested to low-tolerance users is the optimized minimum required to achieve targeted satisfaction.

| PARAMETER NAME | PARAMETER VALUE |
|---|---|
| Number of eNBs | 1 |
| Number of users | 3 |
| Number of available RBs | 9 |
| Number of subcarriers per RB | 12 |
| Resource block bandwidth | 180 kHz |
| Carrier frequency | 2 GHz |
| Flat fading | Rayleigh |
| Log normal shadowing | 8dB standard deviation |
| Distance attenuation | L=35.3+37.6 × log(d) |
| UE thermal noise figure | 9 dB |
| UE thermal noise density | −174 dBm/Hz |
| Grid size (k) | 100 |
| $S_{min}$ | 4 |

Table 1. Simulation parameters.

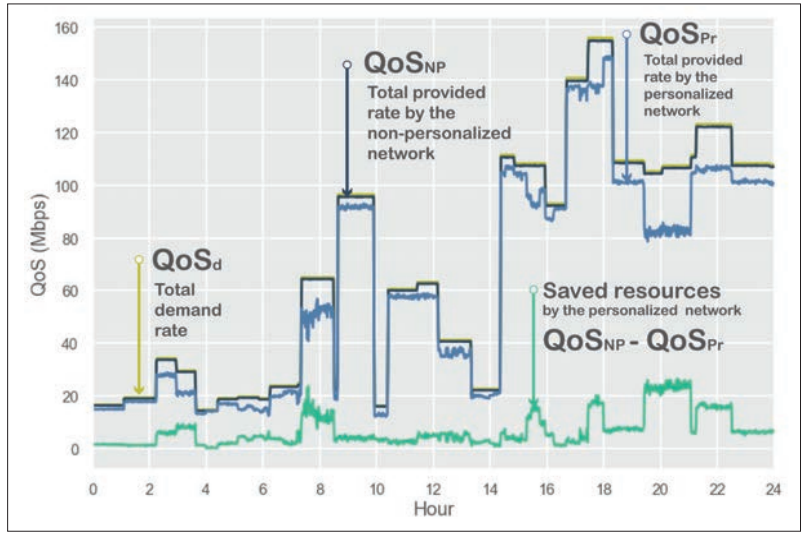

Figure 4. Total $QoS_{NP}$, $QoS_{Pr}$, $QoS_d$, and $QoS_{NP} - QoS_{Pr}$ for the three users vs. time in hours.

The second premise of personalized networks is their ability to maintain targeted satisfaction levels. To substantiate this claim, we compare the satisfaction levels of the personalized and non-personalized networks. In Fig. 5, we plot the average user satisfaction for the three users vs. time in hours for both networks. As shown in Fig. 5, although on average the non-personalized network achieved higher satisfaction levels (an average of 4.87), the personalized network was able to maintain user satisfaction above the targeted level of 4 (an average of 4.31) and save resources simultaneously.

## OPEN ISSUES

While the proposed personalization framework has the potential to achieve the ultimate balance between the conservation of resources and increasing user satisfaction, there are plenty of open issues that need further study prior to practi-cal implementation. Some of these issues include the following.

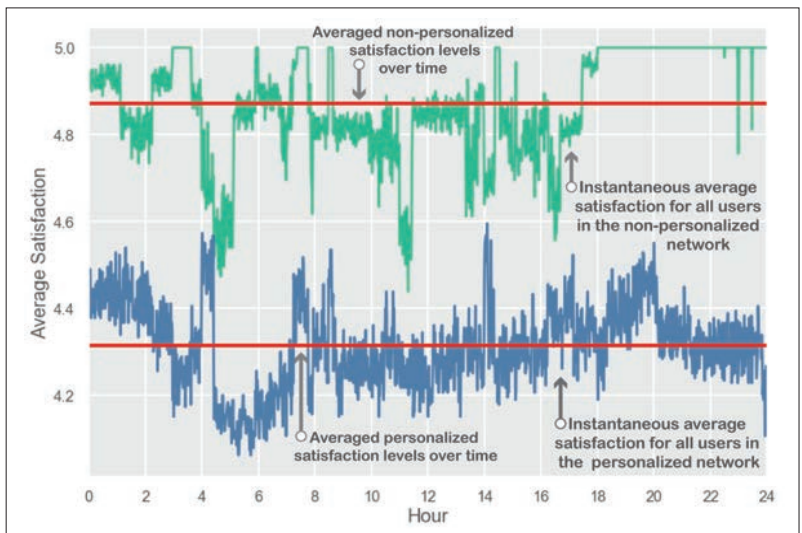

**Figure 5.** Average user satisfaction for the three users vs. time in hours for the personalized and non-personalized networks.

**Dynamic User Data:** User data is highly dynamic and unpredictable. Therefore, the process of data acquisition and preprocessing is extremely challenging, and necessitates more investigation and exploration.

**Non-Intrusive User Feedback:** As mentioned earlier, there are efforts in the literature to sense and predict users' emotions and opinions in a non-intrusive manner using various sensor data. However, the process of correlating the predicted feelings to user satisfaction levels in wireless networks constitutes a challenging but interesting line of research.

**Privacy and Security Issues:** Personalized networks offer attractive means of providing better service quality and value for users along with increased profitability for service providers. This win-win situation, however, is marred by privacy and security challenges since personalizing people's experiences in networks entails gathering considerable amounts of data about them. For this reason, more research should be conducted to reconcile the goals and techniques used for network personalization with privacy and security restrictions.

**User Persona Clustering:** In the early stages of adopting network personalization, networks do not have enough information about the types of personas available. Therefore, unsupervised clustering for user personas is an area of research that needs more study and analysis.

## Conclusions

This article has proposed wireless network personalization as an enabler for resource micro-management based on users' actual demands and needs. Along with AI and big data analytics, personalized networks utilize real-time non-intrusive user feedback coupled with context information to make fine-grained decisions that achieve higher user satisfaction levels using minimum resources. Furthermore, since measuring, tracking, and analyzing user satisfaction is indispensable for personalized networks, we propose the user satisfaction model, which is based on the notion of ZoT. Although this article focuses on resource allocation, personalization can be employed to optimize various decisions in wireless networks, such as network failure detection and network security decisions. Moreover, the technology and framework proposed for wireless networks can be applied to any network with users (e.g., wired networks) as well as other businesses and applications that require user feedback to improve service.

## References

[1] R. Alkurd, I. Abualhaol, and H. Yanikomeroglu, "Dataset Modeling for Data-Driven AI-Based Personalized Wireless Networks," *Proc. 2019 IEEE ICC*, Shanghai, China, May 2019.
[2] R. Alkurd, I. Abualhaol, and H. Yanikomeroglu, , "A Synthetic User Behavior Dataset Design for Data-Driven Aibased Personalized Wireless Networks," *Proc. 2019 IEEE ICC Wksps.*, Shanghai, China, May 2019.
[3] P. Magdalinos *et al.*, "A Context Extraction and Profiling Engine for 5G Network Resource Mapping," *Computer Commun.*, vol. 109, Sept. 2017, pp. 184–201.
[4] R. Schoenen and H. Yanikomeroglu, "User-in-the-Loop: Spatial and Temporal Demand Shaping for Sustainable Wireless Networks," *IEEE Commun. Mag.*, vol. 52, no. 2, Feb. 2014, pp. 196–203.
[5] Y. Bao, H. Wu, and X. Liu, "From Prediction to Action: Improving User Experience with Data-Driven Resource Allocation," *IEEE JSAC*, vol. 35, no. 5, May 2017, pp. 1062–75.
[6] X. Hu *et al.*, "Emotion-Aware Cognitive System in Multi-Channel Cognitive Radio Ad Hoc Networks," *IEEE Commun. Mag.*, vol. 56, no. 4, Apr. 2018, pp. 180–87.
[7] M. S. Hossain *et al.*, "Audio–Visual Emotion-Aware Cloud Gaming Framework," *IEEE Trans. Circuits Sys. Video Tech.*, vol. 25, no. 12, Dec. 2015, pp. 2105–18.
[8] M. S. Hossain and G. Muhammad, "Emotion-Aware Connected Healthcare Big Data Towards 5G," *IEEE Internet of Things J.*, vol. 5, no. 4, Aug. 2018, pp. 2399–2406.
[9] R. Subramanian *et al.*, "Ascertain: Emotion and Personality Recognition Using Commercial Sensors," *IEEE Trans. Affective Computing*, vol. 9, no. 2, Apr. 2018, pp. 147–60.
[10] S. Li and W. Deng, "Reliable Crowdsourcing and Deep Locality Preserving Learning for Unconstrained Facial Expression Recognition," *IEEE Trans. Image Processing*, vol. 28, no. 1, Sept. 2018, pp. 356–70.
[11] A. Parasuraman, V. A. Zeithaml, and L. L. Berry, "A Conceptual Model of Service Quality and Its Implications for Future Research," *J. Marketing*, vol. 49, no. 4, Oct. 1985, pp. 41–50.
[12] A. Kaloxylos, "A Survey and an Analysis of Network Slicing in 5G Networks," *IEEE Commun. Standards Mag.*, vol. 2, no. 1, Mar. 2018, pp. 60–65.
[13] R. Alkurd, I. Abualhaol, and H. Yanikomeroglu, "A Synthetic User Behavior Dataset Design for Data-Driven AI-Based Personalized Wireless Networks"; https://github.com/rawanalkurd/Personalization-Framework-Datasets, accessed Feb. 25, 2020.

## Biographies

RAWAN ALKURD (Rawan.Alkurd.A@ieee.org) is a Ph.D. student in the Department of Systems and Computer Engineering at Carleton University, Ottawa, Canada. In 2016, she received the Vanier graduate scholarship, Canada's most prestigious graduate scholarship. Her research interests include big data, artificial intelligence, machine learning, and their applications in wireless networks.

IBRAHIM ABUALHAOL [SM] (ibrahimee@ieee.org) is a principal data scientist at Huawei Technologies and an adjunct research professor at Carleton University. He holds a B.Sc, an M.Sc, and a Ph.D. in electrical and computer engineering. He also holds an M.Eng in technology innovation management. He is a Professional Engineer (P.Eng.) in Ontario, Canada. His research interests include machine learning and real-time big data analytics and their applications in the Internet of Things, cybersecurity, and wireless communications.

HALIM YANIKOMEROGLU [F] (halim@sce.carleton.ca) is a full professor in the Department of Systems and Computer Engineering at Carleton University. His research interests cover many aspects of wireless technologies with special emphasis on wireless networks. His collaborative research with industry has resulted in 35 granted patents. He is a Fellow of the Engineering Institute of Canada (EIC) and the Canadian Academy of Engineering (CAE). He is a Distinguished Speaker for both the IEEE Communications Society and the IEEE Vehicular Technology Society.